\pdfoutput=1
\documentclass[10pt,logo,copyright]{nvidiatechreport}
\usepackage[authoryear,sort&compress,round]{natbib}

\usepackage[utf8]{inputenc} 
\usepackage[T1]{fontenc}    

\usepackage{parskip}        
\usepackage{url}            
\usepackage{hyperref}       
\usepackage{booktabs}       
\usepackage{amsfonts}       
\usepackage{nicefrac}       
\usepackage{microtype}      
\usepackage{xcolor}         
\usepackage[dvipsnames]{xcolor} 
\usepackage{graphicx}
\usepackage{animate}        
\usepackage{subcaption}
\usepackage{tabularx}
\usepackage{makecell}
\usepackage{adjustbox}
\usepackage{setspace}
\newcolumntype{M}[1]{>{\centering\arraybackslash}m{#1}}
\usepackage{float}
\usepackage{tikz}
\usetikzlibrary{positioning,shapes,arrows}
\usepackage{amsmath,amsfonts,bm, bbm,leftindex}
\usepackage{multirow}
\usepackage{comment}
\usepackage{gensymb}
\usepackage{lipsum}
\usetikzlibrary{arrows.meta, positioning, fit}
\usepackage[para]{threeparttable}
\usepackage{tikz}
\usetikzlibrary{tikzmark}

\def\eqref#1{equation~\ref{#1}}

\def\1{\bm{1}}

\DeclareMathAlphabet{\mathsfit}{\encodingdefault}{\sfdefault}{m}{sl}
\SetMathAlphabet{\mathsfit}{bold}{\encodingdefault}{\sfdefault}{bx}{n}

\makeatletter
\let\save@mathaccent\mathaccent
\newcommand*\if@single[3]{%
  \setbox0\hbox{${\mathaccent"0362{#1}}^H$}%
  \setbox2\hbox{${\mathaccent"0362{\kern0pt#1}}^H$}%
  \ifdim\ht0=\ht2 #3\else #2\fi
  }
\newcommand*\rel@kern[1]{\kern#1\dimexpr\macc@kerna}
\newcommand*\widebar[1]{\@ifnextchar^{{\wide@bar{#1}{0}}}{\wide@bar{#1}{1}}}
\newcommand*\wide@bar[2]{\if@single{#1}{\wide@bar@{#1}{#2}{1}}{\wide@bar@{#1}{#2}{2}}}
\newcommand*\wide@bar@[3]{%
  \begingroup
  \def\mathaccent##1##2{%
    \let\mathaccent\save@mathaccent
    \if#32 \let\macc@nucleus\first@char \fi
    \setbox\z@\hbox{$\macc@style{\macc@nucleus}_{}$}%
    \setbox\tw@\hbox{$\macc@style{\macc@nucleus}{}_{}$}%
    \dimen@\wd\tw@
    \advance\dimen@-\wd\z@
    \divide\dimen@ 3
    \@tempdima\wd\tw@
    \advance\@tempdima-\scriptspace
    \divide\@tempdima 10
    \advance\dimen@-\@tempdima
    \ifdim\dimen@>\z@ \dimen@0pt\fi
    \rel@kern{0.6}\kern-\dimen@
    \if#31
      \overline{\rel@kern{-0.6}\kern\dimen@\macc@nucleus\rel@kern{0.4}\kern\dimen@}%
      \advance\dimen@0.4\dimexpr\macc@kerna
      \let\final@kern#2%
      \ifdim\dimen@<\z@ \let\final@kern1\fi
      \if\final@kern1 \kern-\dimen@\fi
    \else
      \overline{\rel@kern{-0.6}\kern\dimen@#1}%
    \fi
  }%
  \macc@depth\@ne
  \let\math@bgroup\@empty \let\math@egroup\macc@set@skewchar
  \mathsurround\z@ \frozen@everymath{\mathgroup\macc@group\relax}%
  \macc@set@skewchar\relax
  \let\mathaccentV\macc@nested@a
  \if#31
    \macc@nested@a\relax111{#1}%
  \else
    \def\gobble@till@marker##1\endmarker{}%
    \futurelet\first@char\gobble@till@marker#1\endmarker
    \ifcat\noexpand\first@char A\else
      \def\first@char{}%
    \fi
    \macc@nested@a\relax111{\first@char}%
  \fi
  \endgroup
}
\makeatother

\newcommand{\modelname}{LocateAnything3D}

\definecolor{darkred}{rgb}{0.7, 0.0, 0.0}

\usepackage{pifont}
\usepackage{wrapfig}

\usepackage[nameinlink]{cleveref}
\crefname{equation}{Eq.}{Eqs.}
\crefname{figure}{Fig.}{Figs.}
\crefname{section}{Sec.}{Sec.}
\crefname{appendix}{App.}{App.}
\crefname{table}{Tab.}{Tabs.}
\crefname{algorithm}{Algo}{Algo}
\crefname{thm}{Thm}{Thm}
\Crefname{thm}{Thm}{Thm}
\crefname{prop}{Prop}{Prop}
\usepackage{longtable}
\usepackage{afterpage}

\newcommand{\crefnames}[3]{%
  \@for\next:=#1\do{%
    \expandafter\crefname\expandafter{\next}{#2}{#3}%
  }%
}

\title{\modelname{}: Vision-Language 3D Detection with Chain-of-Sight}

\author{Yunze Man$^{1*}$, ~~Shihao Wang$^{2}$, ~~Guowen Zhang$^{2}$, ~~Johan Bjorck, ~~Zhiqi Li, ~~Liang-Yan Gui$^{1}$, ~~Jim Fan, ~~~~~ ~~Jan Kautz,  ~~Yu-Xiong Wang$^{1\dag}$, ~~Zhiding Yu$^\dag$}

\begin{abstract}
    To act in the world, a model must name what it sees and know where it is in 3D. Today's vision-language models (VLMs) excel at open-ended 2D description and grounding, yet multi-object 3D detection remains largely missing from the VLM toolbox. We present \textbf{LocateAnything3D}, a VLM-native recipe that casts 3D detection as a next-token prediction problem. The key is a short, explicit Chain-of-Sight (CoS) sequence that mirrors how human reason from images: find an object in 2D, then infer its distance, size, and pose. The decoder first emits 2D detections as a visual chain-of-thought, then predicts 3D boxes under an easy-to-hard curriculum: across objects, a near-to-far order reduces early ambiguity and matches ego-centric utility; within each object, a center-from-camera, dimensions, and rotation factorization ranks information by stability and learnability. This VLM-native interface preserves open-vocabulary and visual-prompting capability without specialized heads. On the challenging Omni3D benchmark, our model achieves state-of-the-art results, with \textbf{38.90} $\mathrm{AP_{3D}}$, surpassing the previous best by \textbf{+13.98} absolute improvement even when the baseline is given ground-truth 2D boxes. It also generalizes zero-shot to held-out categories with strong robustness. By turning 3D detection into a disciplined next-token problem, LocateAnything3D offers a practical foundation for models to perceive in 3D.
\end{abstract}

\begin{document}

\maketitle

\abscontent

\textbf{Links:} \hspace{2pt}
{
\hypersetup{urlcolor=nvidiagreen}
\href{https://nvlabs.github.io/LocateAnything3D/}{Project Page}
}

\section{Introduction}
\label{sec:intro}

Vision-language models (VLMs) have rapidly advanced open-ended perception in 2D: with a single model and a single decoding interface, they localize, describe, and reason about arbitrary image content across diverse domains~\cite{ferret,qwenvl,li2025eagle2}. Yet one capability has lagged behind: general, multi-object 3D detection directly from monocular images. Existing monocular 3D detectors perform well within narrow domains, but rely on task-specific heads, closed label spaces, and carefully calibrated cameras; they do not inherit the versatility, compositionality, or instruction-following behavior that makes VLMs compelling. Recent work begins to bridge the gap by either coupling specialized 3D heads to open-vocabulary 2D detectors~\cite{detany3d,ov3d}, or by prompting foundation models with auxiliary geometric inputs, but they mostly address single-object grounding or require customized modules that break the simplicity of the VLM paradigm~\cite{cubellm}. In short, we still lack a VLM that can \emph{natively} perceive 3D and produce reliable, multi-object 3D boxes from a single image.

The strong motivation behind teaching VLMs to reason about 3D lies in the next frontier of the embodied intelligence: not just perception, but action. 3D boxes are a compact, metrically meaningful scene state: they connect recognition to interaction, make supervision verifiable, and enable calibration in diverse environments. Folding this capability into the same, token-based interface that already handles 2D grounding simplifies system design and makes scaling with data straightforward. The question we pursue is focused: \emph{what is the most VLM-native recipe that makes multi-object monocular 3D detection just work}?

We answer this question with \emph{Chain-of-Sight (CoS)}, a decoding and supervision scheme that teaches 3D the way humans often reason from pictures: first commit to what is visible in 2D, then infer distance, size, and pose~\cite{marr2010visionbook,von1867handbuch,rock1983logic}. More specifically, we cast detection as a short token sequence that interleaves 2D and 3D per instance: the decoder emits a 2D box, then the corresponding 3D box, and repeats until an end-of-sequence token. The explicit 2D step serves as a high-confidence visual chain of thought that focuses the search on the right pixels, ties subsequent tokens to verifiable evidence, and reduces hallucination. In an autoregressive model, this is not merely convenient formatting: early tokens should be easy, highly informative, and attributable. Committing to image space first provides strong conditioning for the rest of the sequence, shapes the likelihood landscape to be smoother for 3D tokens, and yields a natural interface for prompting. Because the same decoder accepts either text or visual cues, a user can supply text instruction or a box/click, and the model continues with the 3D state for that instance without switching heads or losses.

\begin{figure*}[!t]
    \centering
    \includegraphics[trim=0 0 0 0, clip=True, width=1\textwidth]{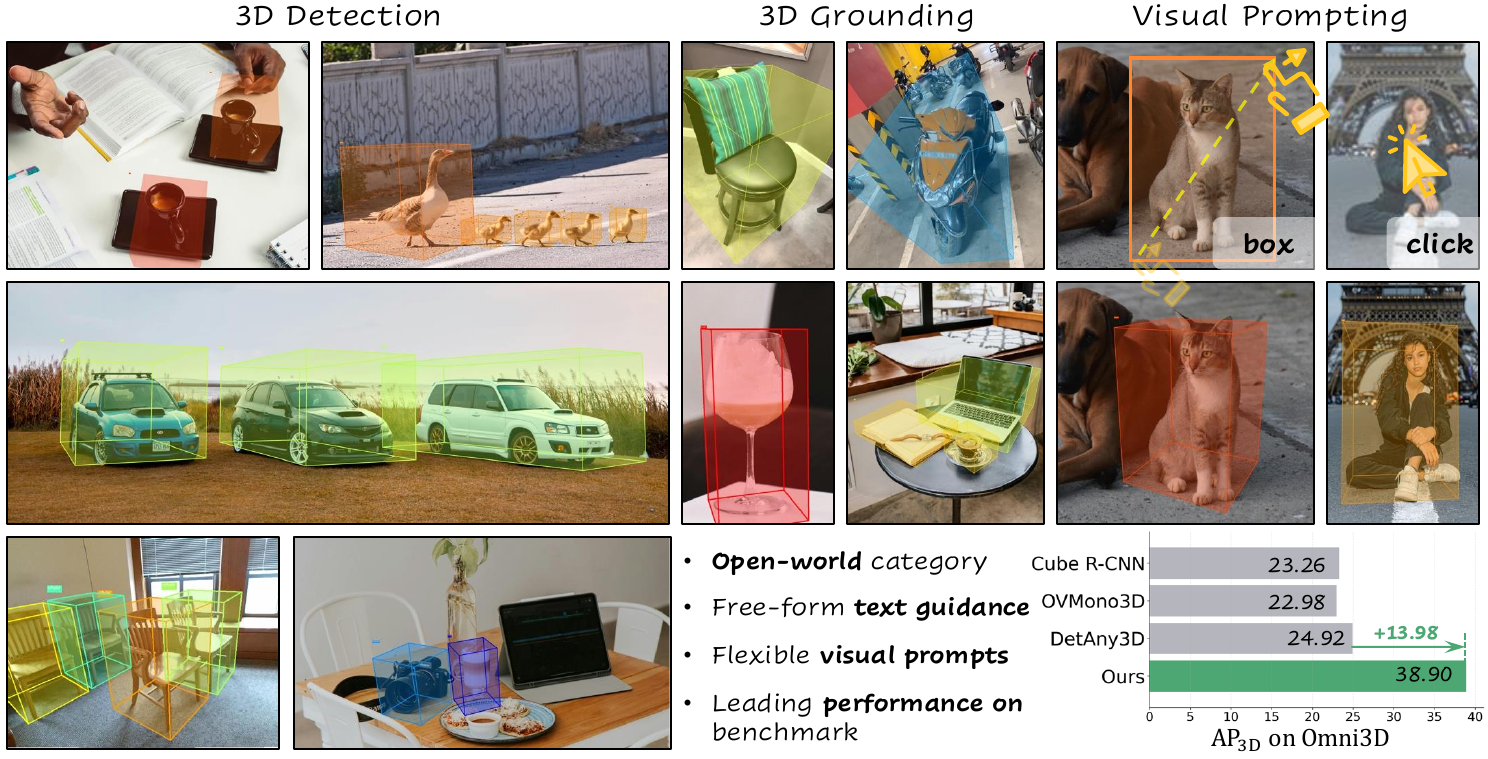}
    \vspace{-6mm}
    \captionof{figure}{\textbf{\modelname{}} unifies 3D detection and grounding in a single vision-language model. It supports open-world categories with free-form text guidance and flexible visual prompts (\eg, drag boxes, click points). All examples are zero-shot, highlighting strong out-of-domain generalizability. The bar chart (right) shows that \modelname{} achieves state-of-the-art $\mathrm{AP_{3D}}$ on Omni3D benchmark.}
    \vspace{-4mm}
    \label{fig:teaser}
\end{figure*}

Beyond the 2D proxy, we align supervision to the natural curriculum of autoregressive decoding. \emph{Across objects}, we serialize detections by depth, from near to far. This ordering matches ego-centric utility (near objects matter first), provides high-evidence tokens early, and sets geometric context that constrains scale and distance for later objects via relative size and occlusion. Placing ambiguous, far instances at the tail prevents them from derailing the prefix. \emph{Within each object}, we factorize the 3D box into a semantically ordered tuple and decode \emph{center} $\rightarrow$ \emph{size} $\rightarrow$ \emph{rotation}. This ranking mirrors the observability of monocular cues: ``where is it?'' before ``how big is it?'' before ``how is it oriented?'', and stabilizes learning by letting location constrain the latter properties. Compared to corner-based encodings that entangle all parameters and amplify early errors, this factorization is both more learnable and better calibrated.

To train CoS end-to-end, we curate a camera-centric corpus that presents supervision in exactly the sequence the model will decode: 2D $\rightarrow$ 3D and near $\rightarrow$ far. We unify heterogeneous data sources into a shared schema, retain intrinsics and a consistent camera-frame parameterization, and convert the data into VLM conversations with calibrated negatives for anti-hallucination. The resulting package is a high-quality dataset that comprises approximately 1.74M training examples spanning indoor and outdoor scenes and diverse camera rigs for 3D vision-language perception.

The results demonstrate the power of our 2D-as-proxy and easy-to-hard curricula. On the challenging Omni3D dataset~\cite{brazil2023omni3d}, our method attains state-of-the-art performance with \textbf{38.90} ${\rm AP_{3D}}$, surpassing the previous best by \textbf{+13.98} absolute points even when the baseline is aided by ground-truth 2D boxes. The same model shows strong zero-shot generalization to held-out categories. Ablations corroborate the design: replacing near-to-far with a left-to-right scanline or random ordering drops performance by a large margin. Removing the 2D CoS also collapses accuracy. Qualitative results (Figs.~\ref{fig:teaser} and~\ref{fig:qualitative}) show depth-consistent ordering, scale stability across repeated objects, and coherent orientations under occlusion and truncation. This paper makes three contributions: 
\begin{itemize}
    \item A \textbf{Chain-of-Sight} formulation that turns open-world monocular 3D detection into a native next-token prediction problem in a VLM. By coupling explicit 2D grounding with 3D decoding, CoS improves reliability while preserving text- or visual-prompting within one interface.
    \item A \textbf{curriculum and representation} tailored to autoregressive decoding: near$\rightarrow$far serialization across objects and an intra-object tokenization that yields consistent decoding, stronger performance and robustness under camera and category shifts.
    \item A \textbf{camera-centric dataset} that unifies heterogeneous data sources into CoS-ready corpus, enabling scalable and systematic ablations without task-specific heads.
\end{itemize}
These elements deliver simple, strong, and broadly applicable 3D perception within a VLM, closing a long-standing gap between open-vocabulary recognition and metric 3D understanding.

\section{\modelname{}}
\label{sec:method}

\begin{figure*}[!t]
    \centering
    \includegraphics[width=\linewidth]{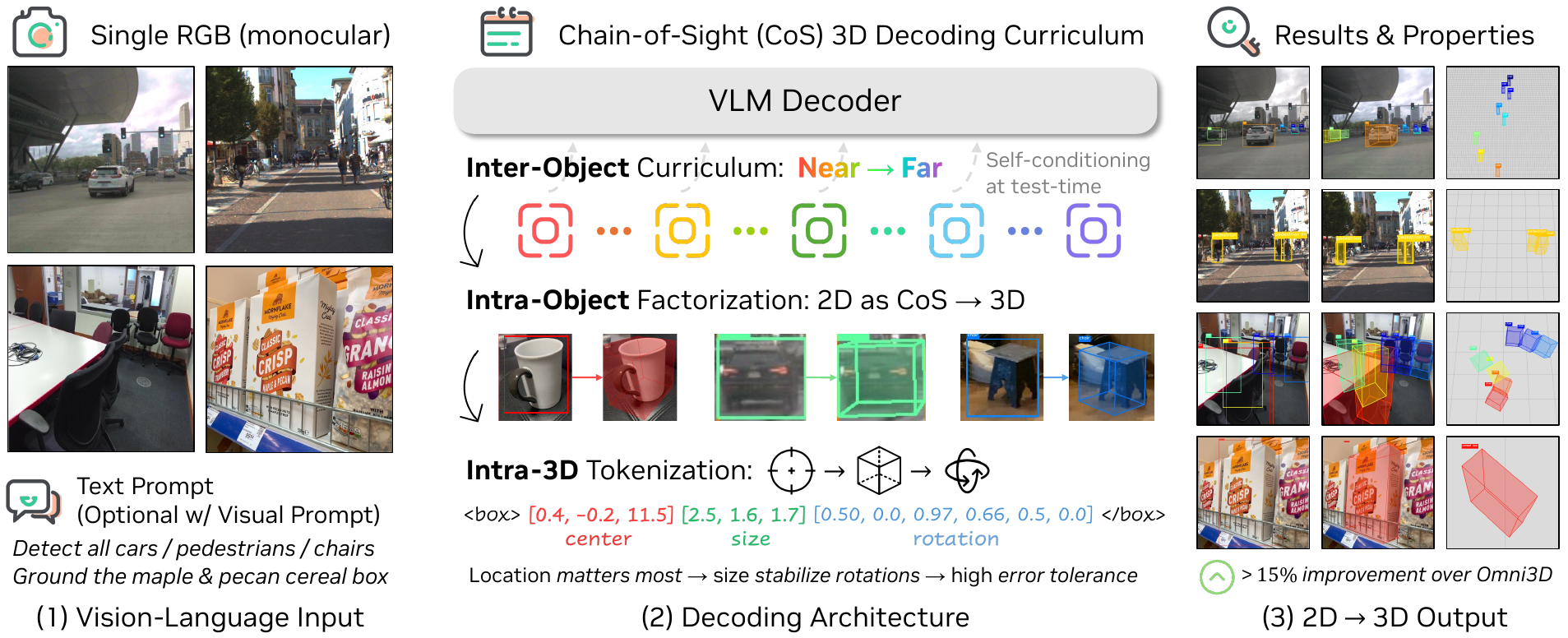}
    \vspace{-6mm}
    \caption{\textbf{Architecture of \modelname{}.} (1) Model input: a single RGB image with text and optional visual prompts (boxes/clicks). (2) Chain-of-Sight (CoS) decoding: a VLM decoder first emits 2D detections as an explicit visual evidence, then continues the sequence to 3D. Decoding follows three layers of design: inter-object curriculum ordering detections from near to far; intra-object factorization using 2D as CoS to robustly infer 3D; and intra-3D tokenization that outputs center, size, and rotation. (3) We output calibrated multi-object 3D boxes with open-vocabulary categories and flexible prompting, yielding strong results on Omni3D. We use turbo colormap for boxes to demonstrate their depth, where \colorbox{red!20}{\textcolor{black}{reddish}} and \colorbox{blue!20}{\textcolor{black}{blueish}} colors indicate closer and farther objects, respectively.}
    \vspace{-4mm}
    \label{fig:architecture}
\end{figure*}

\textbf{Overview.} We study the monocular, open-world 3D detection task in a VLM-native setting, as demonstrated in our architecture diagram in Fig.~\ref{fig:architecture}. A single RGB image and free-form text query drive an autoregressive (AR) decoder that emits a short, structured sequence comprising 2D proposals and their 3D counterparts. The core idea is our \emph{Chain-of-Sight (CoS)} factorization, which makes 3D a native next-token prediction problem. 

\subsection{Preliminaries: Monocular 3D Detection} Let $I\in\mathbb{R}^{H\times W\times 3}$ be a monocular RGB image and let $c\in\Sigma^*$ denote a free-form textual description of a target category (\eg, ``car,'' ``any cup,'' or ``red chair''). The goal is to predict a variable-sized set of 3D bounding boxes of that category, denoted as $\mathcal{B}_c = \{\mathbf{b}_i\}_{i=1}^{N_c}$. We represent a 3D box in the camera coordinate frame as \begin{equation} 
    \mathbf{b}_i = (\mathbf{t}_i, \mathbf{d}_i, \mathbf{R}_i) \quad \mathbf{t}_i\in\mathbb{R}^3; \mathbf{d}_i\in\mathbb{R}^3_{+}; \mathbf{R}_i\in\mathrm{SO}(3), \label{eq:boxrep} 
\end{equation} 
where $\mathbf{t}_i=(X_i,Y_i,Z_i)^\top$ is the 3D center from the camera, $\mathbf{d}_i=(W_i,H_i,L_i)^\top$ are metric dimensions, and $\mathbf{R}_i$ is the object rotation. In scenes that admit the upright-world assumption (\eg, autonomous driving), $\mathbf{R}_i$ can be parameterized by a single yaw angle; our formulation remains valid for the general case.

Given $I$ and $c$, monocular 3D detection can be posed as set inference $\hat{\mathcal{B}}_c = \arg\max_{\mathcal{B}} P(\mathcal{B}| I,c)$ where $P(\mathcal{B}| I,c)$ is the conditional distribution of all 3D boxes of the queried category. With an autoregressive decoder, a standard factorization of the previous equation is \begin{equation} P(\mathcal{B} | I,c) = \prod_{i=1}^{N_c} P(\mathbf{b}_i\mid I, c, \mathbf{b}_{<i}), \label{eq:ar_set} \end{equation} where $\mathbf{b}_{<i}$ denotes previously generated boxes and an end-of-sequence token handles the unknown cardinality $N_c$.

For later use, we also define the 2D bounding box of instance $i$ as $\mathbf{q}_i = (x^\mathrm{min}_i, y^\mathrm{min}_i, x^\mathrm{max}_i, y^\mathrm{max}_i)\in\{0,1,\ldots,1000\}^4$ in normalized integer image coordinates, and a (known or estimated) pinhole projection operator $\Pi$ mapping $(\mathbf{t}_i,\mathbf{d}_i,\mathbf{R}_i)$ to image space. We write $\Pi(\mathbf{b}_i)\Rightarrow \mathbf{q}_i$ when the 2D box is obtained by projecting the 3D cuboid.

\subsection{Chain-of-Sight (CoS) Factorization} The key innovation is to interleave 2D and 3D in the token sequence so that 2D localization acts as a visual chain-of-thought, which we call chain-of-sight (CoS), that constrains 3D inference. Concretely, the decoder emits 
\begin{equation} 
    \mathcal{S} = (\mathbf{q}_1, \mathbf{b}_1, \mathbf{q}_2, \mathbf{b}_2, \ldots, \langle\mathrm{eos}\rangle), \label{eq:sequence} 
\end{equation} 
where each 2D box $\mathbf{q}_i$ is immediately followed by its 3D counterpart $\mathbf{b}_i$. The resulting conditional probability decomposes as 
\begin{equation}
P(\mathcal{S}\mid I,c) = \prod_{i=1}^{N_c} \underbrace{P(\mathbf{q}_i\mid I,c, \mathcal{S}_{<i})}_{\text{2D localization}} \underbrace{P(\mathbf{b}_i\mid I,c, \mathcal{S}_{<i}, \mathbf{q}_i)}_{\text{3D estimation}} \cdot P(\langle\mathrm{eos}\rangle\mid I,c,\mathcal{S}_{\le N_c}), 
\label{eq:cos}
\end{equation}
where $\mathcal{S}_{<i}$ denotes all tokens emitted before step $i$. Compared to Eq.~\eqref{eq:ar_set}, this CoS factorization introduces a high-confident intermediate $\mathbf{q}_i$ that: (i) focuses the search on the right pixels, (ii) reduces hallucination by tying 3D tokens to visible evidence, and (iii) aligns naturally with AR decoding, where early tokens should be both easy and highly informative. By committing to $\mathbf{q}_i$ first, the model learns to ground each instance before decoding its 3D state, mirroring how textual chain-of-thought stabilizes hard reasoning.

\textbf{Inter-object curriculum.} Conventional 2D detectors often impose a scanline or left-to-right ordering when serializing detections for AR decoders~\cite{ferret,li2025eagle2}. Such policies are agnostic to 3D geometry: two boxes that are adjacent in 2D may be at very different depths. Hence, far-away instances that are intrinsically ambiguous in monocular views can appear adjacent and early in the sequence and derail subsequent decoding. We therefore adopt a \emph{near-to-far} curriculum across objects. Placing nearer objects first improves three properties relevant to 3D: (1) \emph{utility}: nearer instances matter most for interaction and safety; (2) \emph{evidence quality}: close objects provide stronger monocular cues, yielding confident early tokens; and (3) \emph{context}: once nearby geometry is established, it constrains the plausible size and depth of distant objects via relative scale and occlusion relationships. In practice, the depth-aware order leads to more stable, well-calibrated sequences than 2D scanline order.

\textbf{Intra-object factorization (2D $\Rightarrow$ 3D).} Although 3D detection does not \emph{require} predicting 2D boxes, we deliberately ask the model to do so. Prior monocular methods commonly rely on an \emph{external} 2D detector to propose boxes and then lift them to 3D with a specialized head~\cite{yao2024open,liu2024grounding,detany3d}. In contrast, our VLM performs both 2D localization and 3D estimation \emph{within the same decoder and the same interface}. This tight coupling is beneficial for the same reason textual chain-of-thought helps language problems: intermediate commitments break a hard prediction into easier, verifiable steps. Here, the 2D prediction serves as a \emph{visual CoT} -- our Chain-of-Sight -- that anchors subsequent 3D tokens. The design also naturally supports visual prompting: when a user supplies a 2D cue (\eg, a box or a click), the decoder can immediately continue with the corresponding 3D tokens for that instance, preserving the AR workflow.

\textbf{Intra-3D tokenization.} A 3D box can be represented in several ways. Corner-based encodings list eight projected or 3D vertices~\cite{brazil2023omni3d}, but they are ambiguous to an AR decoder (which corner comes first?), and amplify early-token errors. Instead, we adopt the structured representation of Eq.~\eqref{eq:boxrep} and, crucially, a \emph{semantic ordering} for AR decoding: {center} $\mathbf{t}$ → {size} $\mathbf{d}$ → {rotation} $\mathbf{R}$. This order reflects information value and difficulty: ``where is it?'' before ``how big is it?'' before ``how is it oriented?'', and we find it substantially improves the robustness.

\textbf{Coordinate and rotation systems.} We predict boxes in the \emph{camera} frame rather than a world frame. This avoids burdening the model with estimating scene-level coordinates (which vary across datasets and camera rigs) and improves cross-domain generalization. Projection to image space uses the usual pinhole model, $\Pi: (\mathbf{t},\mathbf{d},\mathbf{R})\mapsto \mathbf{q}$, with intrinsics known or estimated. For rotation, our formulation supports either a full $\mathrm{SO}(3)$ rotation or a yaw-dominant parameterization when the upright assumption is reasonable (\eg, driving scenes). The latter allocates most capacity to the most observable angle under monocular cues while retaining the general case when needed. Overall, the CoS factorization (Eq.~\ref{eq:cos}), together with a near-to-far inter-object curriculum and center$\rightarrow$size$\rightarrow$rotation intra-object ordering, turns open-world monocular 3D detection into a compact sequence that is easy for a VLM to learn and robust to decode, all within a single, unified interface. Training uses standard cross-entropy losses over tokens; additional details follow in subsequent sections.
\section{LocateAnything3D Data Curation at Scale}

\textbf{Goal.} We construct a large, camera-centric corpus that natively supports our Chain-of-Sight decoding (Fig.~\ref{fig:architecture}) and the formulation in Sec.~\ref{sec:method}. The data are presented to the model exactly in the sequence it will decode at test time: first 2D, then 3D, and from near to far. We unify heterogeneous monocular 3D benchmarks into a single representation and package them as VLM conversations for both single-object grounding and multi-object detection.

\textbf{Datasets and Unification.} We leverage six public 3D detection datasets: {ARKitScenes}~\cite{baruch2021arkitscenes}, {SUN-RGBD}~\cite{song2015sunrgbd}, {Hypersim}~\cite{roberts2021hypersim}, {Objectron}~\cite{objectron2021}, {KITTI}~\cite{geiger2013vision}, and {nuScenes}~\cite{caesar2020nuscenes} into a shared JSONL format. Across datasets we retain camera intrinsics and adopt a camera-coordinate convention for 3D boxes to maximize cross-domain transfer.

\subsection{Stage I: Canonical Multi-Box Normalization}
\textbf{Output unit.} For each image and category we create one JSONL line containing all instances of that category, ordered by depth. Formally, each line corresponds to a tuple $(\texttt{image\_path},\texttt{category\_name})$ and carries a list of per-instance fields aligned by index.

\textbf{Geometry-based filtering and quality control.} We drop instances that are behind the camera or entirely outside the image frustum relative to the camera frame. When the dataset provides visibility and truncation metadata, we keep items with visibility greater than 0.16 and truncation less than 0.84; otherwise we approximate these terms using 2D projections, depth ordering, and border intersection. These thresholds balance coverage and precision, removing ambiguous supervision that is particularly harmful early in autoregressive decoding.

\textbf{2D and 3D representations.} To represent 2D objects, we store both tight pixel boxes $\mathbf{q}=(x^{\min},y^{\min},x^{\max},y^{\max})$ and normalized coordinates in $[0,1000]$ (integers). The 2D representation can also be conveniently converted to center point format for prompting variants (\eg, points). For 3D reprensetation, we keep multiple redundant parameterizations to support ablations and alternative supervision choices for each instance: (i) the center in camera coordinates $\mathbf{t}=(X,Y,Z)$ (meters); (ii) dimensions $\mathbf{d}=(W,H,L)$ (meters); and (iii) rotation as a $3\times3$ matrix $\mathbf{R}$, Euler angles (ZYX) rescaled to $[0,1]$, and their element-wise sine/cosine (mapped from $[-1,1]$ to $[0,1]$). Numeric fields are rounded to two decimals (zero preserved) to control entropy while retaining salient signal. Each line also stores image width/height and the intrinsic matrix $\mathbf{K}$. Within each grouped line we sort by increasing depth of the 3D center from the camera. This stage yields approximately \textbf{480K} single-image, multi-object training entries.

\subsection{Large-Scale Text Auto-Annotation} To supply rich referring expressions without manual labeling, we prompt strong VLMs~\cite{gemini2.5,gpt4v} on images where exactly one target instance is highlighted at a time (a single tight 2D box overlay; the scene remains otherwise untouched). Prompts ask for concise, \emph{unambiguous} descriptions that uniquely identify the target using semantic attributes, spatial layout (left/right/top/bottom; nearby objects), coarse pose, and contextual anchors.

We generate three paraphrases per target with mild sampling for lexical diversity, then conduct automated uniqueness checks: (i) contrastive A/B re-rendering on another instance of the same category; (ii) candidate-index selection tests; and (iii) rejection of hedged or unverifiable language. The resulting corpus contains \textbf{$\sim$1.0M} high-quality single-object grounding samples.

\subsection{Negative Samples for Anti-Hallucination} We explicitly supervise \emph{no-match} behavior. For each image we know the exact set of present categories from the canonical lines. We sample absent categories, including hard negatives chosen via semantic proximity (\eg, car and van), and produce queries that should yield no detections. Negatives are capped at 10\% of training examples (at most 2 per training image), so positives dominate while every batch carries calibrated rejection pressure. Packaging is identical to positives except the model must emit a sentinel token \texttt{<no\_object/>}. This simple design significantly reduces false positives without harming recall.

\subsection{Stage II: Packaging for VLM Training} We convert the canonical JSONL into conversational samples suitable for an autoregressive decoder.

\textbf{Conversation record.} Each example has a unique \texttt{id}, an \texttt{image} pointer, and a two-turn dialogue: a human prompt and a model response. The response concatenates one or more instance segments, each containing a 2D box immediately followed by its 3D counterpart (mirroring CoS). Multi-object examples preserve the near-to-far order inherited from Stage~I.

\textbf{Scale and generalization.} The same processing applies to all datasets. The unified schema allows us to scale training without dataset-specific heads, and enables consistent ablations on ordering, representation choices, and instruction phrasing across all sources. Combining normalized detection entries, single-object grounding, and calibrated negatives yields approximately \textbf{1.74M} training conversations spanning diverse categories, camera rigs, and scene types, which will be made publicly available.
\begin{table*}[tbp] 
    \centering
    \vspace{-2mm}
    \caption{ 
    \textbf{3D detection on the Omni3D benchmark.} Our LocateAnything3D achieves state-of-the-art results over all baselines, even outperform DetAny3D with additional ground-truth 2D inputs on metrics. The first three columns (Omni3D\_{OUT}) show outdoor-only results, while the remaining columns show results on the full unified dataset spanning indoor and outdoor scenes.
    } \label{tab:indomain}
    \vspace{-3mm}
    \resizebox{1\linewidth}{!}{
    \begin{tabular}{l|ccc|ccccccc}
    \toprule
    \multirow{2}{*}{Method} & \multicolumn{3}{c|}{\rm Omni3D\_{OUT}}
    & \multicolumn{7}{c}{\rm Omni3D}\\
   & ${\rm AP^{kit}_{3D}} \uparrow$ 
    & ${\rm AP^{nus}_{3D}} \uparrow$  & ${\rm AP^{out}_{3D}} \uparrow$ 
    & ${\rm AP^{kit}_{3D}} \uparrow$ 
    & ${\rm AP^{nus}_{3D}} \uparrow$ 
    & ${\rm AP^{sun}_{3D}} \uparrow$ 
    & ${\rm AP^{ark}_{3D}} \uparrow$ 
    & ${\rm AP^{obj}_{3D}} \uparrow$ 
    & ${\rm AP^{hyp}_{3D}} \uparrow$ 
    & ${\rm AP_{3D}} \uparrow$ \\
    \midrule
   
    ImVoxelNet~\cite{rukhovich2022imvoxelnet} &  23.5 & 23.4 & 21.5 & -&-&-&-&-&-& 9.4\\
    SMOKE~\cite{liu2020smoke} &  25.9 & 20.4 &20.0 & -& - & - & - & - & - & 10.4  \\
   
     OV-Uni3DETR~\cite{wang2023uni3detr} &35.1& 33.0 & 31.6 &  -& -&- &- &- & - & - \\
    Cube R-CNN~\cite{brazil2023omni3d} & 36.0 & 32.7 &  31.9   & {32.50} & 30.06 & 15.33 & 41.73 & 50.84 & 7.48 & 23.26 \\
    
    \midrule
   
    $\text{OVMono3D~\cite{ov3d}}$ & - &-    &- & 25.45 &  24.33 & 15.20 & 41.60 & 58.87 & {7.75} & 22.98 \\
   {$\text{DetAny3D}$} &  35.8& {33.9} & {32.2} & 31.61 & {30.97}  & {18.96} &   {46.13} & 54.42 & 7.17  & {24.92} \\
   \rowcolor{gray!20}
    {$\text{DetAny3D}_{\text{{\textit{w/}~Ground-Truth 2D Box}}}$}  & 38.0  & \textbf{36.7} & {35.9} &  {38.68} & \textbf{37.55} & \textbf{46.14} & {50.62} & 56.82 & {15.98} &  {34.38 }\\
    \midrule
    \rowcolor{blue!10} \textbf{$\text{LocateAnything3D}$}  & \textbf{39.8} & \underline{33.9} & \textbf{36.1} & \textbf{43.75} & \underline{35.26} & \underline{45.12} & \textbf{59.89} & \textbf{71.90} & \textbf{18.12} & \textbf{38.90} \\
        \bottomrule
    \end{tabular}
    }
    \vspace{-3mm}
\end{table*}

\section{Experiments}
\label{sec:exp}

\textbf{Benchmarks and metrics.} We evaluate on Omni3D~\cite{brazil2023omni3d}, a large-scale monocular 3D detection suite covering both indoor and outdoor imagery. Omni3D provides official trainval and test splits. The test set is held out strictly: no images or labels from test are used during training or hyperparameter tuning. For evaluation metrics, unless otherwise stated, we adopt the benchmark metrics used in Omni3D. Reported scores are 3D Average Precision ($\rm {AP}_{3D}$) computed over a sweep of 3D IoU thresholds
($\tau \in \{0.05,0.10,\ldots,0.50\}$). Intersections are measured volumetrically in the camera frame, consistent with Sec.~\ref{sec:method}. All evaluations follow a \textit{target-aware} protocol, as advocated in prior open-vocabulary works~\cite{detany3d,ov3d}: for each image, the detector is prompted only with the categories that actually occur in its annotations rather than an exhaustive vocabulary. This simple change alleviates naming inconsistencies and focuses the comparison on 3D localization quality rather than on taxonomy alignment.

\textbf{Baselines.} We compare against methods that are most compatible with our open-world, prompt-driven setup: (1) {Cube R-CNN}~\cite{brazil2023omni3d}: the reference baseline released with Omni3D, a unified detector trained as a close-vocabulary model. (2) {OVMono3D}~\cite{ov3d}: an open-vocabulary monocular 3D detector tailored to Omni3D. It ``lifts'' 2D detections to 3D by wiring an open-vocabulary 2D localizer~\cite{groundingdino} to a 3D prediction head. (3) {DetAny3D}~\cite{detany3d}: a promptable monocular 3D detector that accepts category text and outputs 3D boxes directly, designed for open-world settings.

\textbf{Pretraining of 2D detection and grounding.} Before training the full Chain-of-Sight model, we conduct a 2D detection and grounding pretraining phase to equip the model with strong 2D localization capabilities. This stage focuses exclusively on predicting 2D bounding boxes from text or visual prompts, establishing a robust foundation for the subsequent 2D-to-3D learning. After pretraining, we train the complete CoS sequence (2D $\rightarrow$ 3D) end-to-end using standard cross-entropy loss over the autoregressive token sequence. Additional training details, hyperparameters, and ablations are provided in the supplementary material.

\textbf{Implementation details.} Our work is built on SigLIP vision encoder~\cite{siglip} and Qwen2-8B backbone~\cite{qwenvl} coupled by a lightweight MLP projector. Images are decomposed into up to 12 adaptive tiles plus a global thumbnail, each with 448 pixel size, and the resulting visual tokens replace repeated \texttt{<IMG\_CONTEXT>} tokens in a Qwen2-style chat template. We train with bfloat16 and FlashAttention 2 for both vision and language, apply dynamic online packing to fill a 16,384-token context per sample, and optimize with AdamW and a learning rate of 1e-5, a weight decay of 0.05, a cosine scheduler, and a 3\% warm-up, under ZeRO-3 with gradient checkpointing. Training uses 64 H100 GPUs for 46 hours over 37K steps. Please refer to the supplementary material for complete implementation details.

\begin{figure*}[!t]
    \centering
    \includegraphics[width=\linewidth]{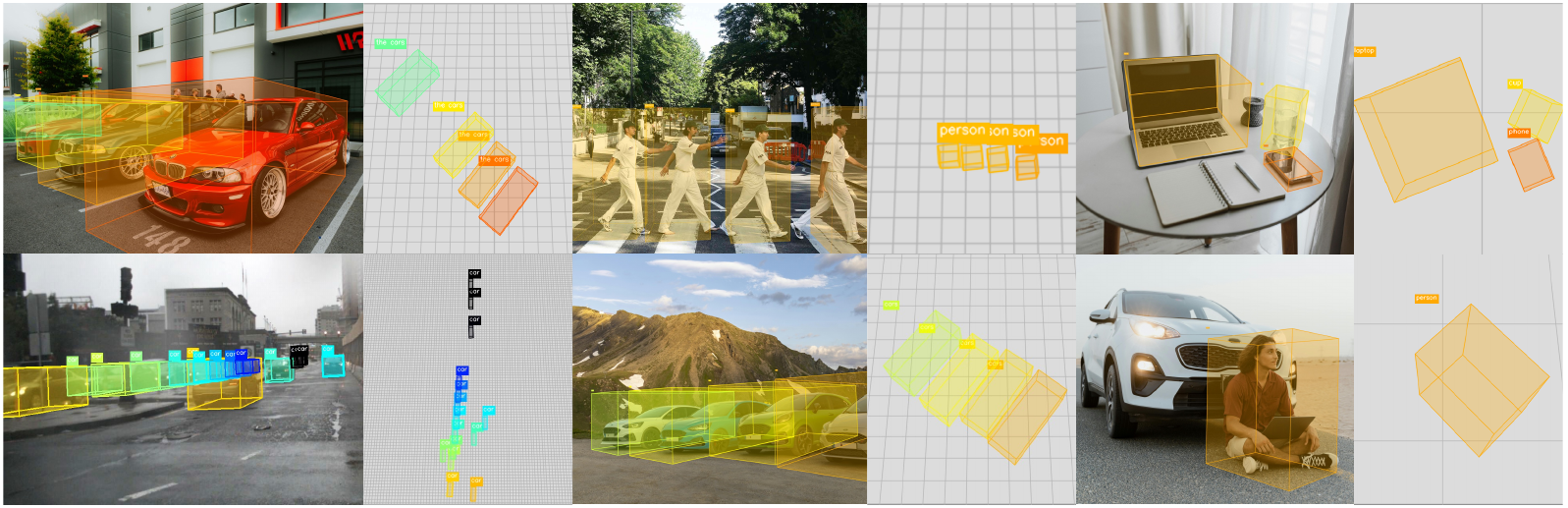}
    \vspace{-6mm}
    \caption{\textbf{Qualitative results of \modelname{}.} For each example, the left sub-figure overlays the projected 3D bounding boxes on the input image, while the right sub-figure shows the corresponding bird's-eye view with 1m$\times$1m grids as the background. We use a turbo colormap based on depth, where \colorbox{red!20}{\textcolor{black}{redish}} colors indicate objects closer to the camera, and \colorbox{blue!20}{\textcolor{black}{blueish}} colors indicate objects farther away.}
    \vspace{-3mm}
    \label{fig:qualitative}
\end{figure*}

\subsection{Main Performance}

\textbf{Overall evaluation and protocal.} Table~\ref{tab:indomain} summarizes results on the Omni3D benchmark. Our \textbf{LocateAnything3D} attains the best score on every metric and every split. On the outdoor-only training/evaluation track (\texttt{Omni3D\_OUT}), our method reaches \textbf{33.1} $\rm AP_{3D}$, overperforming DetAny3D (32.2) and only behind DetAny3D aided by \emph{ground-truth} 2D boxes (35.9). When trained and evaluated on the full unified indoor and outdoor dataset, we also markedly lead across all domains against existing methods, and we even outperform previous method with ground-truth 2D visual prompts in 5 out of 7 metrics. Our overall mean $\rm AP_{3D}$ reaches \textbf{38.90}, surpassing the prior best with privileged 2D boxes by \textbf{+4.52}. These improvements reflect the benefits of CoS decoding: accurate 2D grounding as a first-class step simplifies monocular 3D inference without relying on any auxiliary detectors or oracle boxes.

Following Omni3D, the \texttt{Omni3D\_OUT} setting trains solely on outdoor driving data (KITTI+nuScenes) and reports per-domain and aggregated outdoor metrics; the full Omni3D setting trains on the entire corpus and evaluates across both indoor and outdoor domains. Importantly, all training excludes the official \texttt{test} images and labels. Notably, even methods that receive \emph{ground-truth} 2D boxes at inference lag behind our end-to-end approach, which highlights that learning 2D and 3D jointly within a single autoregressive interface is more effective than bolting a 3D head onto externally supplied 2D proposals.

\begin{wraptable}{R}{0.55\textwidth}
\vspace{-5mm}
\begin{center}
    \caption{LocateAnything3D achieves the best zero-shot 3D detection performance, demonstrating strong generalization to unseen object classes. Notably, baseline methods rely on an external detector for 2D box as additional input, while our method jointly predicts both 2D and 3D boxes end-to-end from a single image alone. Following existing methods, we report $\rm{AP_{3D}}$ using the target-aware metric (per-image existing categories for prompting).}
    \vspace{-3mm}
    \label{tab:novel}
    \resizebox{\linewidth}{!}{
    \begin{tabular}{lcccc}
    \toprule
    \multirow{2}{*}{Method} & \multicolumn{3}{c}{Novel Categories} \\
    \cmidrule(lr){2-4}
      & $\rm{AP^{kit }_{3D}}$ & $\rm{AP^{sun }_{3D}}$ & $\rm{AP^{ark }_{3D}}$ \\
    \midrule
    
    {$\text{OVMono3D}_{\textit{w/}~\text{Grounding-DINO 2D Boxes}}$} & 4.71 & 16.78 & 13.21 \\
    {$\text{DetAny3D}_{\textit{w/}~\text{Grounding-DINO 2D Boxes}}$} & 25.73 & 21.07 & 24.56 \\[2pt]
    \midrule
    \rowcolor{blue!10}
    {$\textbf{LocateAnything3D}_{\text{(single image, no external 2D)}}$} & \textbf{25.87} & \textbf{26.33} & \textbf{29.06} \\
    \cellcolor[rgb]{0.9,0.95,1}\textcolor[rgb]{0,0.3,0.6}{\footnotesize $\Delta$ vs. DetAny3D} & \cellcolor[rgb]{0.9,0.95,1}\textcolor[rgb]{0,0.3,0.6}{\footnotesize +0.14} & \cellcolor[rgb]{0.9,0.95,1}\textcolor[rgb]{0,0.3,0.6}{\footnotesize +5.26} & \cellcolor[rgb]{0.9,0.95,1}\textcolor[rgb]{0,0.3,0.6}{\footnotesize +4.50} \\
    \bottomrule
    \end{tabular}
    }
    \vspace{-4mm}
\end{center}
\vspace{-2mm}
\end{wraptable}

\textbf{Zero-shot novel categories.}\label{sec:novel}
We follow the evaluation protocol used by prior open-vocabulary methods~\cite{ov3d,detany3d}: images are prompted only with the categories present in their annotations, and the held-out classes are never seen during training. As shown in Table~\ref{tab:novel}, \textbf{LocateAnything3D} delivers the strongest zero-shot performance on all benchmarks, with \textbf{25.87} on KITTI novel classes~\cite{geiger2013vision}, \textbf{26.33} on SUN-RGBD~\cite{song2015sunrgbd}, and \textbf{29.06} on ARKitScenes~\cite{baruch2021arkitscenes}, while competing approaches depend on an external 2D detector (Grounding DINO~\cite{groundingdino}) to supply proposals. Relative to DetAny3D+2D, we gain (+0.14), (+5.26), and (+4.50) points on the three metrics, respectively; compared to OVMono3D+2D, our margins widen further. These results support our motivation that predicting 2D and 3D \emph{together} -- rather than lifting from external 2D -- improves transfer to unseen categories.

\subsection{Quantitative Evaluation of 3D Grounding}

\begin{table*}[!t]
    \centering
    \caption{\textbf{Indoor 3D Object Grounding Performance.} We compare LocateAnything3D against Cube-LLM trained on different data scales. $\text{Cube-LLM}_{\text{small}}$ is trained on the LV3D-small subset, while $\text{Cube-LLM}_{\text{large}}$ is trained on the full LV3D dataset containing approximately \textbf{9.6M images}. In contrast, our model is trained on a much smaller curated dataset of \textbf{1.7M images}. Despite this significant disparity in data scale, LocateAnything3D outperforms the best baseline by a large margin across all benchmarks. We report Average Precision ($\text{AP}$) prompted with either category names ($\rm AP^{{cat}}_{3D}$) or category plus spatial location ($\rm AP^{{cat+loc}}_{3D}$).}
    \vspace{-2mm} 
    \label{tab:indoor_grounding}
    \resizebox{0.75\textwidth}{!}{
    \begin{tabular}{lccccccc}
    \toprule
    \multirow{2}{*}{Method} & \multicolumn{2}{c}{Objectron} & \multicolumn{2}{c}{ARKitScenes} & \multicolumn{2}{c}{SUN-RGBD} \\
    \cmidrule(lr){2-3} \cmidrule(lr){4-5} \cmidrule(lr){6-7}
      & $\rm AP^{{cat}}_{3D}$ & $\rm AP^{{cat+loc}}_{3D}$ & $\rm AP^{{cat}}_{3D}$ & $\rm AP^{{cat+loc}}_{3D}$ & $\rm AP^{{cat}}_{3D}$ & $\rm AP^{{cat+loc}}_{3D}$ \\
    \midrule
    $\text{Cube-LLM}_{\text{small}}$~\cite{cubellm} & 56.7 & 36.1 & 21.6 & 28.3 & 25.5 & 25.5 \\
    $\text{Cube-LLM}_{\text{large}}~\cite{cubellm}$ & 69.8 & 45.4 & 23.5 & 31.8 & 29.7 & 28.8 \\
    \midrule
    \rowcolor{blue!10}
    \textbf{LocateAnything3D (Ours)} & \textbf{72.5} & \textbf{75.0} & \textbf{41.7} & \textbf{53.9} & \textbf{29.7} & \textbf{39.5} \\
    \cellcolor[rgb]{0.9,0.95,1}\textcolor[rgb]{0,0.3,0.6}{\footnotesize $\Delta$ vs. $\text{Cube-LLM}_{\text{large}}$} & 
    \cellcolor[rgb]{0.9,0.95,1}\textcolor[rgb]{0,0.3,0.6}{\footnotesize +2.7} & 
    \cellcolor[rgb]{0.9,0.95,1}\textcolor[rgb]{0,0.3,0.6}{\footnotesize +29.6} & 
    \cellcolor[rgb]{0.9,0.95,1}\textcolor[rgb]{0,0.3,0.6}{\footnotesize +18.2} & 
    \cellcolor[rgb]{0.9,0.95,1}\textcolor[rgb]{0,0.3,0.6}{\footnotesize +22.1} & 
    \cellcolor[rgb]{0.9,0.95,1}\textcolor[rgb]{0,0.3,0.6}{\footnotesize +0} & 
    \cellcolor[rgb]{0.9,0.95,1}\textcolor[rgb]{0,0.3,0.6}{\footnotesize +10.7} \\
    \bottomrule
    \end{tabular}
    }
\end{table*}

\textbf{Problem setting.} To further evaluate our LocateAnything3D's capability in following spatial language instructions, we conduct experiments on indoor 3D grounding benchmarks. We strictly follow the experimental protocol established by Cube-LLM~\cite{cubellm}. Specifically, we repurpose the test sets of three standard indoor detection datasets: \textit{Objectron}~\cite{objectron2021}, \textit{ARKitScenes}~\cite{baruch2021arkitscenes}, and \textit{SUN-RGBD}~\cite{song2015sunrgbd} into grounding benchmarks. The task requires the model to localize particular objects based on text prompts that vary in specificity: (1) Category-only: The prompt contains only the object class name (\eg, ``chair''); and (2) Category+Location: The prompt includes the class name augmented with spatial descriptions derived from the object's position relative to the camera (\eg, ``chair on the left'', ``bookshelf close to camera''). The spatial qualifiers (left/right/center and close/medium/far) are generated based on the 2D image coordinates and depth thresholds defined in the baseline setting. We report the Average Precision ($\rm AP_{3D}$) averaged over $\rm IoU_{3D}$ thresholds of $\tau \in \{0.15, 0.25, 0.50\}$. If multiple objects match the text description, the maximum IoU among them is used for evaluation.

\textbf{Evaluation results.} Table~\ref{tab:indoor_grounding} summarizes the results on the benchmark. We copy the Cube-LLM numbers for models pre-trained on the ``LV3D-small'' and full ``LV3D'' corpora from their paper~\cite{cubellm}, and add our LocateAnything3D model, which is trained using the Chain-of-Sight formulation on our unified 3D corpus. Across all three datasets and both metrics, LocateAnything3D substantially outperforms Cube-LLM, despite no task-specific architecture changes for indoor scenes.

From the table, we can also notice that Cube-LLM~\cite{cubellm} achieves lower performance for $\rm AP^{{cat+loc}}_{3D}$ than $\rm AP^{{cat}}_{3D}$, in two out of the three evaluation scenarios. On the contrary, \modelname{}{} achieves consistent performance improvement when location information is provided to the model as additional conditions. This different clearly highlights the higher capability of our model to interpret spatial descriptions and 3D understanding.

\textbf{Problem with point-cloud grounding benchmarks.} Existing indoor 3D grounding datasets such as ScanRefer~\cite{scanrefer} and ReferIt3D~\cite{referit3d} are explicitly built around point clouds rather than images, and are therefore ill-suited to our monocular 3D detection setting. Each scene in these benchmarks is represented by a single reconstructed point cloud but is associated with many RGB views that only partially observe the scene. Referring expressions are written to identify objects in the global 3D scene, not in a particular camera view, and a single object may be visible in multiple images with very different appearances and levels of occlusion. As a result, there is no unambiguous way to assign a unique image and 3D box pair to each language query, and any attempt to project the point-cloud annotations into 2D would depend on arbitrary choices of viewpoint and visibility thresholds. For this reason, we follow Cube-LLM~\cite{cubellm} and evaluate our model on indoor benchmarks derived from Objectron~\cite{objectron2021}, ARKitScenes~\cite{baruch2021arkitscenes}, and SUN-RGBD~\cite{song2015sunrgbd}, where each image already comes with camera-specific 3D boxes and thus naturally supports monocular 3D detection and grounding.

\subsection{Analysis and Ablations}\label{sec:analysis}

In this section, we conduct ablation study to verify the design choices of our chain-of-sight learning paradigm.

\begin{wraptable}{R}{0.55\textwidth}
\vspace{-5mm}
\begin{center}
    \caption{\textbf{Ablation study of Chain-of-Sight (CoS) design choices.} We evaluate each component of our three-layer decoding design on Omni3D\_{OUT}. All results are reported using $\rm{AP^{out}_{3D}}$. Our full design (highlighted) achieves the best performance, validating the importance of each design choice.}
    \vspace{-3mm}
    \label{tab:ablation}
    \resizebox{\linewidth}{!}{
    \begin{tabular}{l|l|c}
    \toprule
    \textbf{Design Component} & \textbf{Variant} & $\rm{AP^{out}_{3D}} \uparrow$ \\
    \midrule
    \multirow{3}{*}{\textbf{Inter-Object Curriculum}} 
    & Random Ordering & 17.5 \\
    & Left-to-Right Ordering & 30.6 \\
    & \cellcolor{blue!10}\textbf{Near-to-Far Ordering} & \cellcolor{blue!10}\textbf{33.1} \\
    \midrule
    \multirow{3}{*}{\textbf{Intra-Object Factorization}} 
    & No 2D (Direct 3D) & 22.7 \\
    & 3D-then-2D & 26.2 \\
    & \cellcolor{blue!10}\textbf{2D-then-3D (CoS)} & \cellcolor{blue!10}\textbf{33.1} \\
    \midrule
    \multirow{3}{*}{\textbf{Intra-3D Tokenization}} 
    & Rotation-Size-Center & 28.8 \\
    & Center-Rotation-Size & 32.9 \\
    & \cellcolor{blue!10}\textbf{Center-Size-Rotation} & \cellcolor{blue!10}\textbf{33.1} \\
    \bottomrule
    \end{tabular}
    }
    \vspace{-4mm}
\end{center}
\vspace{-2mm}
\end{wraptable}

\textbf{Inter-object ordering.} Replacing our depth-aware near $\rightarrow$ far order with common alternatives degrades quality (Table~\ref{tab:ablation}). A random order performs worst (17.5), confirming that sequence position carries semantic load in AR decoding. A left-to-right scanline policy is better (30.6) but still inferior to our near-to-far curriculum (\textbf{33.1}), indicating that 3D-aware serialization (easy, high-evidence instances first) yields more stable and informative token prefixes.

\textbf{Intra-object factorization.} Removing the 2D step and predicting 3D directly drops performance to 22.7. Emitting 3D before 2D (``3D-then-2D'') recovers some accuracy (26.2) but remains far from our CoS layout (\textbf{33.1}). These trends validate the role of 2D as a visual chain-of-thought: committing to image-space evidence makes the subsequent 3D tokens both easier to learn and better calibrated. And it shows that 2D is a helpful signal to learn, even when predicted after the 3D signal. 

\textbf{Intra-3D token order.} Within each object, decoding with the {center}, {size}, and {rotation} ordering performs best. Switching to Rotation-Size-Center harms results (32.9), and Center-Rotation-Size is slightly worse (28.8), suggesting that anchoring location, then scale, before resolving orientation is the most learnable and robust schedule for monocular cues. The small but consistent gap between CSR and CRS further indicates that deferring rotation until after size stabilizes the pose estimate.

\subsection{Data Efficiency and Training Dynamics}

\begin{wrapfigure}{R}{0.55\textwidth}
\vspace{-1mm}
    \includegraphics[width=\linewidth]{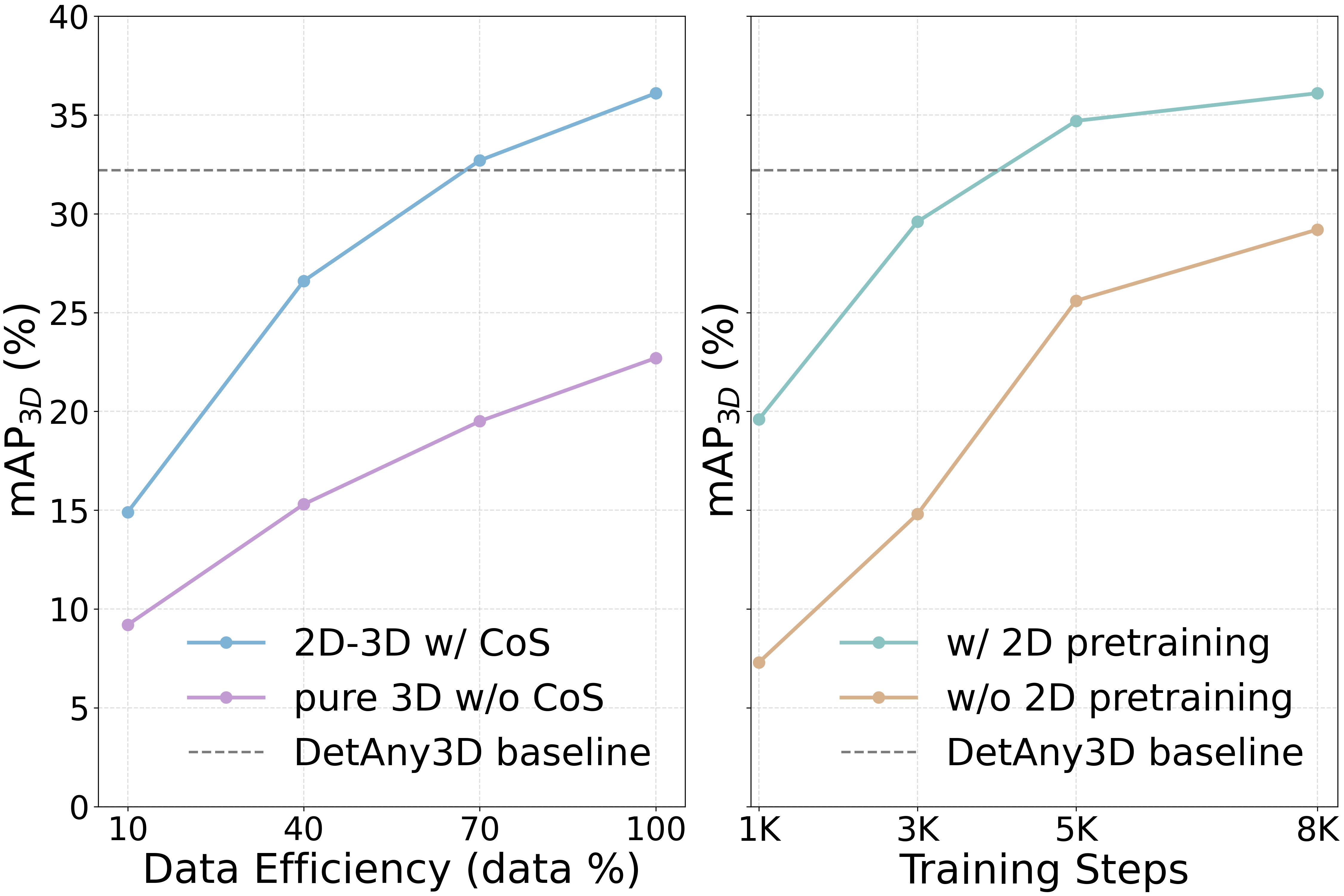}
    \vspace{-6mm}
    \caption{\textbf{Data efficiency and training dynamics analysis}. (1) The left figure shows data efficiency: We report $\rm AP_{3D}$ vs. percentage of training data used. Our Chain-of-Sight (CoS) formulation (\textcolor{blue}{blue}) consistently outperforms direct 3D prediction (\textcolor{violet}{purple}), achieving competitive performance with only 10\% of the data. (2) The right figure shows training dynamics: We compare training curves with and without 2D detection pretraining. 2D pretraining (\textcolor{nvidiagreen}{green}) accelerates convergence significantly, surpassing the previous state of the art (dashed line) almost immediately, whereas training from scratch (\textcolor{orange}{orange}) is slower and yields lower final accuracy.}
    \vspace{-3mm}
\vspace{-2mm}
\label{fig:supp:curves}
\end{wrapfigure}

To better understand the contributions of our Chain-of-Sight (CoS) formulation and the role of 2D pretraining, we conduct a detailed analysis of our model's performance under limited data regimes and different initialization strategies. Figure~\ref{fig:supp:curves} visualizes these comparisons.

\textbf{Impact of chain-of-sight on data efficiency.} Figure~\ref{fig:supp:curves} (left) compares our full 2D-3D CoS formulation against a ``pure 3D'' decoder trained without any explicit 2D step. On the horizontal axis we vary the fraction of our 3D training corpus from $10\%$ to $100\%$; the dashed line marks the performance of DetAny3D~\cite{detany3d} (32.2 $\rm AP_{3D}$).

Across all data regimes, the CoS model is consistently stronger and markedly more data-efficient than the pure 3D variant. With only $10\%$ of the data, CoS outperforms pure 3D prediction baseline by a large margin. As we further scale the data to $70\%$ and $100\%$, the CoS curve continues to climb to 32.7 and 36.1 $\rm AP_{3D}$, whereas pure 3D saturates at 19.5 and 22.7. This supports our central claim that explicitly factorizing 3D detection into a 2D grounding step followed by 3D lifting is not just more accurate, but also significantly more sample-efficient.

\textbf{Impact of 2D pretraining on convergence.} Figure~\ref{fig:supp:curves} (right) studies the effect of the 2D grounding pretraining stage on our training dynamics. We plot $\rm AP_{3D}$ as a function of CoS training steps, comparing models initialized with and without 2D pretraining, and again mark the DetAny3D performance with a dashed line.

Initializing from the 2D grounding stage yields a substantial head start. After only 1k CoS steps, the pretrained model already achieves 19.6 $\rm AP_{3D}$, whereas the model trained from scratch is still at 7.3. As training proceeds, both curves improve, but the gap persists. At the final checkpoint, the model with 2D pretraining converges to 36.1 $\rm AP_{3D}$, while the scratch model lags behind at 29.2. This indicates that robust 2D localization capabilities serve as a critical foundation for 3D perception, allowing the model to focus its capacity on lifting 2D features to 3D space rather than learning basic localization from scratch.

\subsection{Qualitative Results}\label{sec:qual} Figure~\ref{fig:qualitative} showcases representative predictions of \modelname{}. In each example, the left panel overlays the projected 3D cuboids on the RGB frame, while the right panel renders a bird's-eye view with 1m$\times$1m grids. The turbo colormap encodes depth, revealing a depth-consistent ordering that mirrors our CoS decoding: near instances are resolved first and anchor the subsequent geometry. The model handles moderate occlusion and truncation, maintains scale consistency across repeated objects, and preserves orientation structure even at distance.

\section{Related Work}
\label{sec:related}

We situate our contributions at the intersection of three converging directions: VLMs for visual perception that couple recognition with fine-grained grounding; embodied vision beyond static understanding toward spatial reasoning and acting; and 3D object detection from closed-set training to unified open-vocabulary formulations. 
\subsection{Vision-Language Models in Visual Perception}

The 2D perception task provides a perfect playground for vision-language models (VLMs) to learn localizing and reasoning. Classical pipelines rely on specialized ``vision experts'' for recognition and grounding, including contrastive pretraining for open-set recognition and matching~\cite{radford2021clip}, image-text pretraining for retrieval~\cite{li2022blip}, and strong detectors for region-level grounding~\cite{groundingdino,kirillov2023segany,ren2024grounded}. Building atop these capabilities, recent VLM systems either orchestrate experts as tools under a multimodal controller~\cite{liu2023llavaplus,wu2023visual} or pursue unified backbones that natively tackle a wide spectrum of perception tasks~\cite{ferret,xiao2024florence,wang2023cogvlm,wang2024emu3,team2024chameleon,llava,gpt4v}. Within grounding, the long-standing line of referring-expression comprehension (REC) frames 2D localization from unstructured language, with RefCOCO/+/g, Flickr30k-Entities, and subsequent datasets pushing object-level grounding in everyday scenes~\cite{kazemzadeh2014referitgame,yu2016modeling,mao2016generation,plummer2015flickr30k,liu2024finecops,wang2024unveiling}. Recent VLMs further include grounding into training objectives, learning to output boxes or points directly -- \eg, bounding-box supervision in Kosmos-2, Qwen-VL, and Gemini~\cite{peng2023kosmos2,qwenvl,gemini2.0,gemini2.5}, and point-based localization in MoLMo and RoboPoint~\cite{molmo,yuan2024robopoint}. 
Beyond static REC, task-conditioned and temporally aware grounding has emerged as a key frontier for embodied use, including task-driven pointing and procedure-aware grounding~\cite{point-it-out}, and 2D grounding as reasoning chain-of-thought~\cite{shao2024visualcot,argus}. Our formulation adopts this 2D-first perspective: we leverage 2D detections as explicit intermediate tokens to structure perception before lifting to full 3D inference, aligning with evidence that tightly coupling grounding with reasoning yields more reliable downstream behavior~\cite{qwenvl,li2025eagle2}.

\subsection{Vision-Language Models for Embodied Vision}

Foundation models are increasingly leveraged as embodied agents that perceive, reason, and act in long-horizon tasks. Early systems primarily relied on prompting to elicit planning behaviors from VLM backbones~\cite{singh2022progpromptgeneratingsituatedrobot, song2023llmplanner, hu2023lookleapunveilingpower, kim2024contextawareplanningenvironmentawarememory, shin2025socraticplannerselfqabasedzeroshot}, with code and API-centric tool interfaces further improving reliability~\cite{liang2023codepolicieslanguagemodel, silver2024generalized}. Subsequent work introduces supervised finetuning, yielding compact yet capable agents for manipulation~\cite{zawalski2024roboticcontrol, kim2024openvla, lee2025molmoact, liu2025towards, lu2025vla, huang2025tactilevlaunlockingvisionlanguageactionmodels, zhang2025up, zhao2025cot} and household or procedural reasoning~\cite{wu2023embodied, chen2024robogptintelligentagentmaking, ji2025robobrainunifiedbrainmodel}. In parallel, spatial intelligence has emerged as essential competencies for open-world embodiment, with lines of work targeting distance/metric understanding and counting~\cite{yang2024thinking, song2024robospatial, du2024embspatial, zhou2025towards, liao2024reasoning, fu2024blink, chen2024spatialvlm, cai2024spatialbot}, as well as benchmark suites that synthesize complex 3D scenes and tasks~\cite{ray2024sat, cheng2024spatialrgpt}. Embodied pointing and grounding further connect perception to action~\cite{yuan2024robopoint, hong20233d, li2024topviewrs}, and recent efforts augment spatial reasoning with structured reasoning~\cite{yuan2025seeingdoingbridgingreasoning, liu2025spatialcot}. Integrated frameworks exemplify the trend toward generalist embodied VLMs and unify perception, reasoning, and planning at scale~\cite{RoboBrain1, RoboBrain2}, while curated benchmarks continue to expand the supervision landscape~\cite{robopoint, pixmo, pointarena, vsi, mmsi, egoplanbench, egoplanbench2, geminirobotics, cosmosr1, vebrain, embodiedov}. Beyond supervision fine-tuning (SFT), reinforcement-driven training also starts to revolutionize the role of reasoning traces in embodiment~\cite{EmbodiedR1,ERA,Vlaser}. Our work is synergistic with these directions: we target \emph{multi-object 3D perception} as a VLM-native next-token problem by structuring 3D detection with intermediate 2D reasoning and curriculum design. Our formulation provides an explicit, language-aligned perception interface that can plug into embodied agents.

\subsection{3D Object Detection}

Classical monocular 3D object detection has been driven by single-dataset optimization on benchmarks, yielding strong in-domain performance with task-specific architectures but limited robustness under distribution~\cite{geiger2013vision,caesar2020nuscenes,chen2016monocular,wang2021fcos3d,liu2020smoke,zhang2023monodetr,li2024bevformer,huang2022monodtr,wang2022probabilistic,zhou2019objectspoints,zhou2021iafainstanceawarefeatureaggregation,chen2021monorunmonocular3dobject}. Parallel lines of work study multi-sensor fusion and spatio-temporal reasoning to boost accuracy, yet typically inherit closed-set label space constraints~\cite{liang2022bevfusion,lin2022sparse4d,bevguide}. To reduce dataset and camera bias, Omni3D unifies diverse sources and introduces Cube R-CNN, showing that multi-dataset training improves cross-scene generalization for monocular detectors~\cite{brazil2023omni3d}. Subsequent efforts further explore bird-eye's-view formulations across indoor and outdoor settings~\cite{li2024unimode,jhang2025vmind}.
Moving beyond closed vocabularies, open-vocabulary 3D detection seeks to recognize and localize categories beyond those seen during training. Much of the early progress assumes point clouds as input or supervision~\cite{lu2022open,lu2023open,zhang2022pointclip,zhu2023object2scene,cao2024coda,cao2024collaborative,zhang2024fm,zhang2025opensight,peng2025global,russakovsky2015imagenet,zhou2022detecting,wang2024ov}. Closer to our setting, OVMono3D lifts open-vocabulary 2D detections (\eg, from Grounding DINO) into 3D with a unified head~\cite{yao2024open,liu2024grounding}. DetAny3D proposes a promptable 3D foundation model that transfers knowledge from 2D foundation models to monocular 3D via feature aggregation~\cite{detany3d}. We pursue a VLM-native decoding that treats multi-object 3D detection as disciplined next-token inference, leveraging explicit 2D-to-3D factorization to improve generalization across categories and camera configurations.

\section{Conclusion}\label{sec:conclusion} 
We present LocateAnything3D, a VLM-native framework that turns monocular 3D detection into a concise next-token task via \emph{Chain-of-Sight} decoding. By committing to 2D localization before 3D, ordering objects near-to-far, and factorizing each box as center, size, and rotation, our approach aligns supervision with the natural curriculum of autoregressive models. Coupled with a CoS-conformant, multi-domain corpus and a simple training recipe, the method delivers state-of-the-art results on Omni3D, both in-domain and zero-shot to novel categories. We believe that our CoS principle provides a practical route for scaling 3D perception within general-purpose VLMs and opens the door to future extensions in video, multi-view reasoning, and embodied planning. 

\clearpage
\appendix
\section{Erratum}
\label{sec:erratum}
We would like to acknowledge an incorrect implementation of the evaluation in the initial report, which has been corrected in this revision. A root cause of the mistake was that mAP was computed with cross-image matches, instead of being computed on a per-image basis in standard protocol. The updated results are lower than those in the initial report, but the claims and conclusions remain unchanged. Problem definition, model training, inference, and visualization are not affected.\\

\section{Acknowledgments}
\label{sec:ack}

The team would like to thank the valuable discussions and input from Tianyi Xiong, Shaokun Zhang, Guo Chen, Di Zhang, Guilin Liu, Xiaolong Li, Paris Zhang, Yilin Zhao, Subhashree Radhakrishnan, Sifei Liu, Hongxu (Danny) Yin, Valts Blukis, Jonathan Tremblay, Bowen Wen, Yan Chang, Wei Liu, Yan Wang. We would also like to acknowledge the following teams: Metropolis, VILA, LPR Robotics, GEAR Lab, AV Research, and ISSAC Robotics. We would also like to thank the NVIDIA infrastructure team for their prompt and helpful assistance.

\section{Additional Experiments and Analysis}

\subsection{Impact of Token Serialization Strategy}

\begin{wraptable}{R}{0.55\textwidth}
\vspace{-5mm}
\begin{center}
    \caption{\textbf{Ablation of Token Serialization Strategy.} We compare our default \textit{Interleaved} Chain-of-Sight strategy (\rm $2D_i \to 3D_i$) against a \textit{Clustered} strategy where all 2D boxes are predicted before all 3D boxes ($\rm 2D_{1...N} \to 3D_{1...N}$) with average precision ($\rm AP_{3D}$). Models are trained for 1 epoch. The results show that the interleaved strategy is significantly more robust, especially in cluttered scenes where associating separated 2D and 3D sequences becomes difficult.}
    \vspace{-3mm}
    \label{tab:serialization}
    \resizebox{\linewidth}{!}{
    \begin{tabular}{lccc}
    \toprule
    \multirow{2}{*}{Serialization Strategy} & \multicolumn{3}{c}{Average Precision ($\rm AP_{3D}$)} \\
    \cmidrule(lr){2-4}
      & Objectron & KITTI & Hypersim \\
    \midrule
    Clustered ($\rm 2D_{1...N} \to 3D_{1...N}$) & 61.5 & 17.4 & 4.7 \\
    \rowcolor{blue!10}
    \textbf{Interleaved ($\rm 2D_i \to 3D_i$, Ours)} & \textbf{63.0} & \textbf{22.1} & \textbf{11.2} \\
    \midrule
    \textit{Performance Gap} & +1.5  & +4.7 & {+6.5}  \\
    \bottomrule
    \end{tabular}
    }
    \vspace{-3mm}
\end{center}
\vspace{-2mm}
\end{wraptable}

To further validate our Chain-of-Sight (CoS) design of interleaving 2D and 3D tokens on the per-object level ($\rm 2D_i \rightarrow 3D_i$), we compare it against a ``clustered'' decoding strategy. In the clustered setting, the model is trained to predict all 2D bounding boxes for the scene first, followed by all corresponding 3D bounding boxes ($\rm 2D_{1...N} \to 3D_{1...N}$). This ablation tests whether the tight coupling of 2D visual evidence with its corresponding 3D geometry is necessary, or if the model can simply learn two separate phases of detection. We report results trained for 1 epoch on three distinct datasets to analyze performance across different scene complexities. As shown in Table~\ref{tab:serialization}, our interleaved default setting consistently outperforms the clustered strategy. The magnitude of this performance gap is strongly correlated with scene clutter and object density.

\textbf{Object-centric scenes.} On Objectron, which typically contains only 1 or 2 prominent objects per image, the performance gap is minimal ($61.5$ vs. $63.0$). The additional effort for the model to associate the $i$-th 3D box with the $i$-th 2D box is negligible.

\textbf{Structured outdoor scenes.} KITTI scenes contain more objects with large depth range, but they follow a structured distribution (cars on a road) with relatively clear depth ordering. While the gap widens, the model can still maintain reasonable 2D-3D association in the clustered setting.

\textbf{Highly cluttered scenes.} The most significant drop occurs on Hypersim which is characterized by chaotic indoor scenes with dozens of objects and frequent occlusions. In these scenarios, the clustered strategy fails catastrophically. The model struggles to maintain the implicit alignment between the $k$-th 2D box generated early in the sequence and the $k$-th 3D box generated much later, resulting in a big difference between the two setting.

\subsection{Runtime Analysis}
\label{sec:supp:runtime}

Although \modelname{} is primarily designed as a general 3D perception VLM rather than a real-time perception system, we report its end-to-end inference latency for completeness. On average, processing a single image-query pair with LocateAnything3D takes \textbf{683 ms} under our evaluation setup with a single H100 GPU. This wall-clock time consists of three main components: (1) vision encoding of the input image, (2) LLM pre-filling with the textual prompt, and (3) autoregressive generation of the mixed 2D/3D box tokens produced by the Chain-of-Sight decoder.

To isolate the cost of the Chain-of-Sight factorization, we compare our full 2D-3D CoS model with a pure-3D variant that directly predicts 3D boxes without emitting intermediate 2D boxes. Introducing the 2D step increases the average latency by only \textbf{121 ms} (from roughly 562 ms to 683 ms), yet enables the substantial accuracy gains and data-efficiency improvements, as reported in the Section 5 of the main paper. In other words, CoS adds a modest computational overhead while making 3D detection both easier to learn and significantly more accurate.

We emphasize that LocateAnything3D is not meant to replace highly optimized real-time detectors used in latency-critical loops (\eg, onboard obstacle avoidance). Instead, our goal is to endow a general-purpose VLM with strong 3D grounding capabilities so that it can serve as a foundation for downstream tasks such as offline planning, scene understanding, and multimodal agent reasoning. In this context, a sub-second per-image latency is well within an acceptable range, especially given the unified interface and performance benefits brought by the Chain-of-Sight formulation.

\section{Implementation Details}
\label{sec:impl}
\subsection{Models, Tokenization, and Prompting}

\textbf{Model designs.} (1) vision encoder. We use SigLIP~\cite{siglip} with FlashAttention 2~\cite{dao2022flashattention} enabled. (2) Language model. We use a Qwen2 8B causal LM~\cite{qwen2.5} with FlashAttention 2, trained end-to-end (no freezing). (3) Multimodal connector. We use an MLP projector, which maps SigLIP tokens to the LLM hidden space with two-layer MLP. 

\textbf{Image tokenization.} A tiling-based tokenization where we decompose images into patches of a forced image size of 448. The total image tokens scale linearly with the number of tiles.

\textbf{Conversation format.} Qwen2-chat template. Image tokens are inserted by replacing each \texttt{<image>} placeholder with \texttt{<IMG\_START>} followed by repeated \texttt{<IMG\_CONTEXT>} tokens and \texttt{<IMG\_END>}. The repeat count equals per-tile tokens times the number of tiles for that image; we assert a strict match between precomputed and actual counts.

\textbf{Labels.} Only assistant spans are supervised; all instruction tokens are masked. Truncation safety checks keep training targets valid.

\subsection{Dynamic Tiling and Packing}

\textbf{Tiling and image processing.} Images are decomposed into an adaptive grid of 448-pixel tiles, min 1 and max 12 tiles, plus an optional global thumbnail. Tiling policy selects the closest aspect ratio while favoring large area coverage for stability.

\textbf{Sequence construction and online packing.} Our context length is 16,384 tokens per sample. We enable online packing to concatenate multiple short samples until the context budget is filled while tracking sub-sample boundaries in the attention mask. A dummy image is inserted only if the entire packed sample is text-only. Position ids respect packed boundaries; the model supports sequence parallel groups but we run with degree 1 in our experiments.

\begin{table*}[t]\scriptsize
\renewcommand{\arraystretch}{1.2}
    \setlength\tabcolsep{30.4pt}
    \resizebox{\textwidth}{!}{
    \begin{tabular}{m{3cm}|m{14cm}}
\footnotesize Category & \footnotesize Dataset  \\
\hline
\rowcolor{gray!15}\footnotesize 
Captioning \& Knowledge & ShareGPT4o~\cite{opengvlab_sharegpt4o_dataset}, KVQA~\cite{shah2019kvqa}, Movie-Posters~\cite{skvarre_movie_posters_100k}, Google-Landmark~\cite{weyand2020googlelandmark}, WikiArt~\cite{wikiart_dataset}, Weather-QA~\cite{ma2024weatherqa}, Coco-Colors~\cite{mscoco-controlnet-canny-less-colors}, music-sheet~\cite{sheet_music_clean}, SPARK~\cite{yu2024spark}, Image-Textualization~\cite{pi2024image_textualization}, SAM-Caption~\cite{pixart_alpha_sam_llava_captions10m}, Tmdb-Celeb-10k~\cite{ashraq_tmdb_celeb_10k}\\
\footnotesize 
Mathematics & GeoQA+~\cite{cao2022geoqa_plus}, MathQA~\cite{yu2023mathqa}, CLEVR-Math/Super~\cite{lindstrom2022clevrmath, li2023superclevr}, Geometry3K~\cite{lu2021geometry3k}, MAVIS-math-rule-geo~\cite{zhang2024mavis}, MAVIS-math-metagen~\cite{zhang2024mavis}, InterGPS~\cite{lu2021intergps}, Raven~\cite{zhang2019raven}, GEOS~\cite{seo2015geos}, UniGeo~\cite{chen2022unigeo}\\
\rowcolor{gray!15}\footnotesize 
Science &AI2D~\cite{kembhavi2016ai2d}, ScienceQA~\cite{lu2022scienceqa}, TQA~\cite{kembhavi2017tqa}, PathVQA~\cite{he2020pathvqa}, SciQA~\cite{auer2023sciqa}, Textbooks-QA, VQA-RAD~\cite{lau2018vqarad}, VisualWebInstruct~\cite{tiger_lab_visualwebinstruct}
\\\footnotesize 
Chart \& Table &ChartQA~\cite{masry2022chartqa}, MMC-Inst~\cite{liu2023mmcinst}, DVQA~\cite{kafle2018dvqa}, PlotQA~\cite{methani2020plotqa}, LRV-Instruction~\cite{liu2023lrv-instruction}, TabMWP~\cite{lu2022tablemwp}, UniChart~\cite{masry2023unichart}, Vistext~\cite{tang2023vistext}, TAT-DQA~\cite{zhu2022tatdqa}, VQAonBD~\cite{VQAonDB}, FigureQA~\cite{kahou2017figureqa}, Chart2Text~\cite{kantharaj2022chart2text}, RobuT-\{Wikisql, SQA, WTQ\}~\cite{zhao2023robut}, MultiHiertt~\cite{zhao2022multihiertt}\\
\rowcolor{gray!15}\footnotesize 
Naive OCR & SynthDoG~\cite{kim2022synthdog}, MTWI~\cite{he2018icpr2018_MTWI}, LVST~\cite{sun2019lsvt}, SROIE~\cite{huang2019icdar_sroie}, FUNSD~\cite{jaume2019funsd}, Latex-Formula~\cite{oleehyo_latex_formulas}, IAM~\cite{marti2002iam}, Handwriting-Latex~\cite{aida}, ArT~\cite{chng2019art}, CTW~\cite{yuan2019ctw}, ReCTs~\cite{zhang2019rects}, COCO-Text~\cite{veit2016cocotext}, SVRD~\cite{yu2023icdar_svrd}, Hiertext~\cite{long2023icdar_hiertext}, RoadText~\cite{tom2023icdar_roadtext}, MapText~\cite{li2024icdar_maptext}, CAPTCHA~\cite{captcha}, Est-VQA~\cite{wang2020estvqa}, HME-100K~\cite{tal}, TAL-OCR-ENG~\cite{tal}, TAL-HW-MATH~\cite{tal}, IMGUR5K~\cite{krishnan2023textstylebrush_Imgur5K}, ORAND-CAR~\cite{diem2014icfhr_RAND_CAR}, Invoices-and-Receipts-OCR~\cite{mychen76_invoices_receipts_ocr_v1},  Chrome-Writting~\cite{mouchere2016icfhr2016_chrome_writing}, IIIT5k~\cite{mishra2012scene_iiit5k}, K12-Printing~\cite{tal}, Memotion~\cite{ramamoorthy2022memotion}, Arxiv2Markdown, Handwritten-Mathematical-Expression~\cite{Azu}, WordArt~\cite{xie2022toward_wordart}, 
RenderedText~\cite{wendlerc_renderedtext}, Handwriting-Forms~\cite{ift_handwriting_forms}\\
\footnotesize 
OCR QA &  DocVQA~\cite{clark2017docqa}, InfoVQA~\cite{mathew2022infographicvqa}, TextVQA~\cite{singh2019textvqa}, ArxivQA~\cite{li2024multimodal_arxivQA},
ScreencQA~\cite{hsiao2022screenqa}, DocReason~\cite{mplug_docreason25k}, Ureader~\cite{ye2023ureader}, FinanceQA~\cite{Sujet-Finance-QA-Vision-100k}, DocMatrix~\cite{laurenccon2024building_docmatrix}, A-OKVQA~\cite{schwenk2022aokvqa}, Diagram-Image-To-Text~\cite{kamizuru00_diagram_image_to_text}, MapQA~\cite{chang2022mapqa}, OCRVQA~\cite{mishra2019ocrvqa}, ST-VQA~\cite{biten2019stvqa}, SlideVQA~\cite{tanaka2023slidevqa}, PDF-VQA~\cite{ding2023PDFvqa}, SQuAD-VQA, VQA-CD~\cite{mahamoud2024chic_vqa_cd}, Block-Diagram~\cite{shreyanshu09_block_diagram}, MTVQA~\cite{tang2024mtvqa}, ColPali~\cite{faysse2024colpali}, BenthamQA~\cite{mathew2021asking_benthamqa}\\
\rowcolor{gray!15}\footnotesize 
General VQA & LLaVA-150K~\cite{liu2023llava}, LVIS-Instruct4V~\cite{wang2023lvisinstruct4v}, ALLaVA~\cite{chen2024allava},  Laion-GPT4V~\cite{laion_gpt4v_dataset}, LLAVAR~\cite{zhang2023llavar}, SketchyVQA~\cite{tu2023many}, VizWiz~\cite{gurari2018vizwiz}, IDK~\cite{cha2024visually}, AlfworldGPT, LNQA~\cite{pont2020connecting_lnqa}, Face-Emotion~\cite{fastjob_visual_emotional_analysis}, SpatialSense~\cite{yang2019spatialsense}, Indoor-QA~\cite{keremberke_indoor_scene_classification}, Places365~\cite{zhou2017places365}, MMinstruct~\cite{liu2024mminstruct}, DriveLM~\cite{sima2023drivelm}, YesBut~\cite{nandy2024yesbut}, WildVision~\cite{lu2024wildvision}, LLaVA-Critic-113k~\cite{xiong2024llava_critic}, RLAIF-V~\cite{yu2024rlaif_v}, VQAv2~\cite{goyal2017vqav2}, MMRA~\cite{wu2024mmra}, KONIQ~\cite{hosu2020koniq}, MMDU~\cite{liu2024mmdu}, Spot-The-Diff~\cite{jhamtani2018learning_spotthediff}, Hateful-Memes~\cite{kiela2020hateful_memes}, COCO-QA~\cite{ren2015exploring_cocoqa}, NLVR~\cite{suhr2017corpus_nlvr2}, Mimic-CGD~\cite{laurenccon2024matters_mimic_cgd}, Datikz~\cite{belouadi2023automatikz_datikz},
Chinese-Meme~\cite{emo_visual_data_chinese_meme}, IconQA~\cite{lu2021iconqa}, Websight~\cite{laurenccon2024unlocking_websight}\\
\footnotesize 
Text-only & Orca~\cite{lian2023openorca}, Orca-Math~\cite{mitra2024orca}, OpenCodeInterpreter~\cite{zheng2024opencodeinterpreter}
MathInstruct~\cite{yue2023mammoth_mathinstruct}, WizardLM~\cite{xu2023wizardlm}, TheoremQA~\cite{chen2023theoremqa}, OpenHermes2.5~\cite{OpenHermes2_5}, NuminaMath-CoT~\cite{numina_math_datasets}, Python-Code-25k~\cite{flytech_python_codes_25k}, Infinity-Instruct~\cite{baai_infinity_instruct},
Python-Code-Instructions-18k-Alpaca~\cite{iamtarun_python_code_instructions_18k_alpaca}, Ruozhiba~\cite{looksjuicy_ruozhiba}, InfinityMATH~\cite{zhang2024infinitymath}, StepDPO~\cite{lai2024stepDPO}, TableLLM~\cite{zhang2024tablellm}, UltraInteract-sft~\cite{yuan2024advancing_ultrainteract}\\
\rowcolor{gray!15}\footnotesize 
2D Grounding \& Counting & RefCOCO/+/g (en)~\cite{yu2016refcoco,mao2016refcocog}, Objects365~\cite{shao2019objects365}, COCO~\cite{lin2014microsoft}, EgoObjects~\cite{zhu2023egoobjects}, BLIP3-OCR~\cite{blip3-xgenmm}, BDD100K~\cite{bdd100k}, Nuimages~\cite{nuscenes2019}, Flick30K~\cite{plummer2015flickr30k}, LVIS~\cite{gupta2019lvis}
\end{tabular}}
\vspace{-2mm}
\caption{\textbf{Summary of our extensive and diverse supervised fine-tuning datasets for 2D pretraining.} We use a comprehensive collection of numerous large-scale datasets spanning multiple domains and tasks to pretrain our model, ensuring broad coverage and robust performance across diverse visual and language understanding scenarios.}
\label{tab:data_sft}
\end{table*}

\subsection{Optimization and Systems}

We use a precision of bfloat16 across vision and language. For memory handling, gradient checkpointing is enabled for both the SigLIP encoder and the LLM; fused ops are used to reduce memory overhead. For the loss function, we fuse the linear cross-entropy with per-sample normalization using the number of valid answer tokens. For the optimizer and schedule, we use AdamW with a learning rate of 1e-5, a weight decay of 0.05, a cosine decay, and a warm-up of 3\%.

\textbf{Packing target.} We use dynamic online packing to saturate the 16K context; the scripts set an iteration-level token target of $2^{17}$ ($=128K$) to govern accumulation and throughput.

\textbf{Training scale.} We train our model using 64 H100 GPUs. The whole training takes 46 hours with 37K steps, distributed with torchrun and DeepSpeed ZeRO-3.

\subsection{2D Grounding Pretraining}

\textbf{Dataset composition.} We pretrain on a large-scale 2D grounding corpus covering four domains with different data mixture percentage: (1) \textbf{General detection:} Object365~\cite{shao2019objects365} (5 epochs), COCO~\cite{lin2014microsoft} (12 epochs), and LVIS~\cite{gupta2019lvis} (3 epochs); (2) \textbf{Ego-centric \& driving:} BDD100K~\cite{bdd100k} (3 epochs), nuImages~\cite{caesar2020nuscenes} (3 epochs), and EgoObjects~\cite{zhu2023egoobjects} (3 epochs); (3) \textbf{Referring-expression grounding:} RefCOCO~\cite{yu2016refcoco} (3 epochs), RefCOCO+~\cite{mao2016refcocog} (3 epochs), RefCOCOg~\cite{mao2016refcocog} (3 epochs), and Flickr30k~\cite{plummer2015flickr30k} (3 epochs); (4) \textbf{Text grounding:} a BLIP3-OCR subset~\cite{blip3-xgenmm} ($\approx 1.0$M samples). Overall, this results in over \textbf{15M} multi-turn dialogues in the grounding corpus, which we mix with an additional \textbf{8M} samples for general instruction tuning, as demonstrated in Table~\ref{tab:data_sft}.

\textbf{Annotation format.} For each image, we construct a multi-turn dialogue where each turn follows the instruction template ``Detect all the objects in the image that belong to the category set \{c\}.'' The response is either a comma-separated list of 2D bounding boxes in $[x_1,y_1,x_2,y_2]$ format (top-left to bottom-right, integer-quantized to $[0,1000]$), or ``None'' if no instance exists. We include all positive categories present in the image and sample 10 absent categories as negatives, yielding per-image supervision that mixes existence and non-existence signals across multiple dialogue turns.

\section{Limitations and Future Work}

While LocateAnything3D establishes a strong foundation for VLM-native 3D perception, several avenues remain for future exploration. Our work primarily focuses on validating the Chain-of-Sight (CoS) decoding mechanism within a single-frame, end-to-end setting. Below, we outline key directions where our framework can be naturally extended to incorporate additional geometric signals and temporal contexts.

\textbf{Integration of explicit depth priors.} Currently, our model infers metric depth solely from monocular RGB cues and semantic context. While the near-to-far curriculum effectively regularizes this process, the model does not yet leverage explicit depth maps. Future work could introduce a depth encoder or use depth images as an additional conditioning input. This would allow the model to utilize output from state-of-the-art monocular depth estimators as a geometric prompt, potentially improving metric accuracy in texture-less or ambiguous scenes.

\textbf{Explicit camera intrinsic conditioning.} Our current approach normalizes 3D coordinates into a unified camera-centric space to maximize cross-dataset generalization. However, it implicitly relies on the vision encoder to handle variations in focal length and field of view. An extension is to explicitly tokenize camera intrinsic matrices (\eg, focal length, principal point) and feed them as positional prompts. This would allow the decoder to mathematically adjust its size and depth predictions based on the specific camera optics, rather than learning an average projection model.

\textbf{Extension to multi-frame and video settings.} The current framework operates on single images. However, the autoregressive nature of our decoder is naturally suited for temporal sequences. Future iterations could extend the context window to include visual tokens from preceding frames. The model could learn to track objects over time, estimate velocity, and leverage multi-view consistency to resolve depth ambiguities that are ambiguous in a single frame.

\section{Broader Impact}

The development of LocateAnything3D represents a step toward unifying semantic understanding and metric perception within general-purpose foundation models. By enabling VLMs to perceive the physical world in 3D without specialized heads, we lower the barrier to entry for developing capable embodied agents and home robotics. This has positive implications for industries ranging from autonomous driving to assistive robotics.

However, we acknowledge potential risks associated with this technology. Like all deep learning models trained on web-scale data, our model may inherit biases present in the training corpora, such as geographic or cultural biases in object distributions. This could lead to uneven performance across different regions. We encourage the research community to prioritize the development of diverse, representative 3D datasets and to consider the ethical implications of spatial intelligence in deployment scenarios.

\begin{figure*}[!t]
    \centering
    \includegraphics[width=\linewidth]{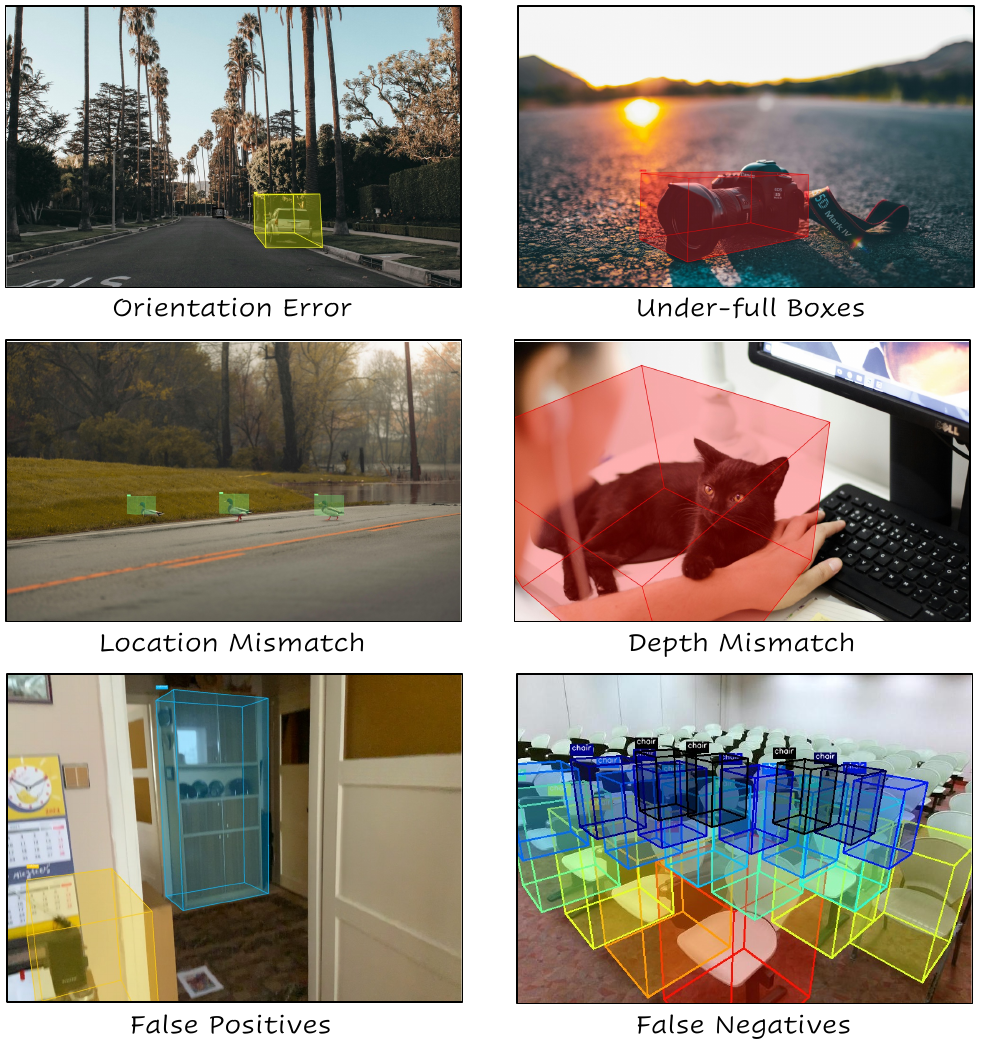}
    \vspace{-4mm}
    \caption{\textbf{Visualization of failure cases}. We show several failure cases of our model. Due to the lack of diverse 3D annotations, similar to the baselines~\cite{detany3d,ov3d}, our model faces challenges when presented with scenes that exhibit very different focal length, spatial layouts, and textural details.}
    \vspace{20mm}
    \label{fig:supp:failure}
\end{figure*}

\begin{figure*}[!t]
    \centering
    \includegraphics[width=\linewidth]{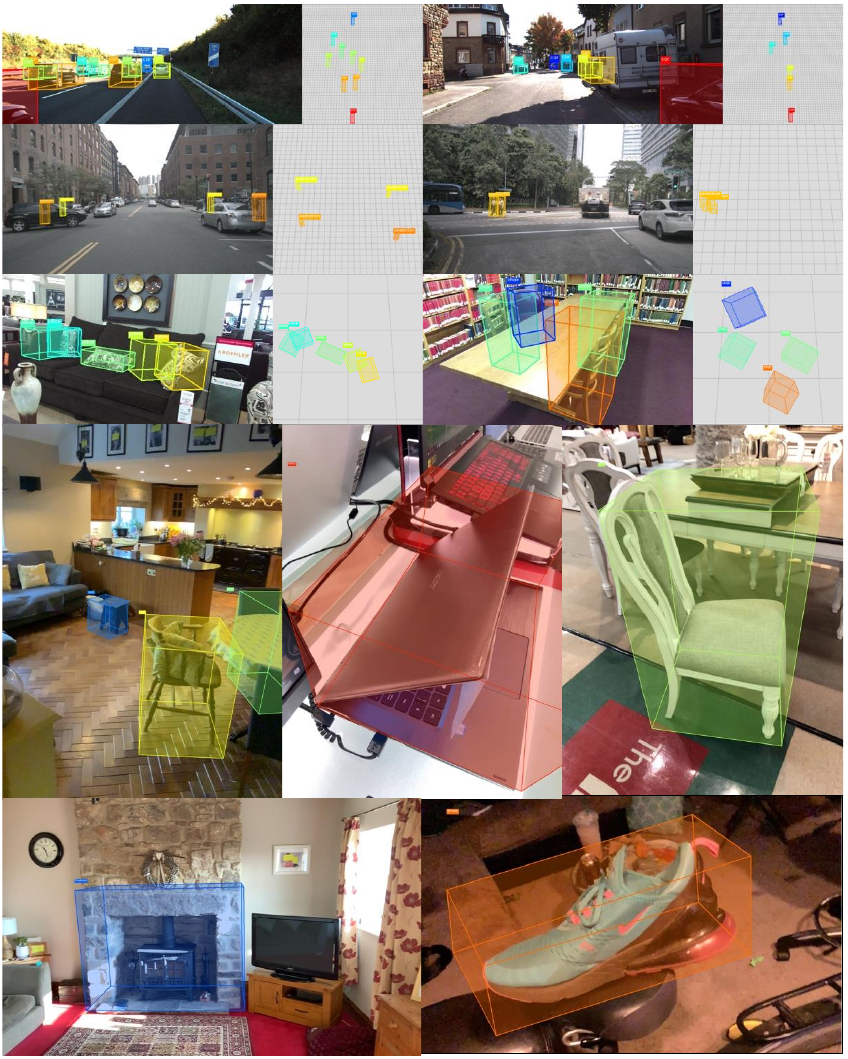}
    \vspace{-7mm}
    \caption{Visualization of more indoor and outdoor successful cases.}
    \vspace{-4mm}
    \label{fig:supp:successful}
\end{figure*}

\section{More Case Visualization}

We provided more qualitative visualization in this section.

\textbf{Failure case visualization.} Figure~\ref{fig:supp:failure} demonstrates some representative failure cases of our method. Despite the great performance, our model still suffers from the lack of diverse and high-quality 3D annotations compared to the 2D scenario. Hence, similar to the baselines~\cite{detany3d,ov3d}, it faces challenges when presented with scenes that exhibit very different focal length, spatial layouts, and textural details.   

\textbf{More successful visualization.} Figure~\ref{fig:supp:successful} demonstrates more successful cases.

\clearpage
\setcitestyle{numbers}
\bibliographystyle{plainnat}
\bibliography{main}

\end{document}